\documentclass{article}

\usepackage{microtype}
\usepackage{graphicx}
\usepackage{subfigure}
\usepackage{booktabs} 

\usepackage{hyperref}

\usepackage[accepted]{icml2025}

\usepackage{amsmath}
\usepackage{amssymb}
\usepackage{mathtools}
\usepackage{amsthm}

\usepackage[capitalize,noabbrev]{cleveref}

\theoremstyle{plain}

\theoremstyle{definition}

\theoremstyle{remark}

\usepackage{enumitem}
\usepackage{pgfplots,tikz}
\pgfplotsset{compat=1.18}
\usetikzlibrary{positioning,calc,arrows.meta}
\definecolor{accent0}{HTML}{1F77B4} 

\setlist[itemize]{nosep,leftmargin=*}
\setlist[enumerate]{nosep,leftmargin=*}

\icmltitlerunning{Video Self-Distillation for Single-Image Encoders}

\begin{document}

\twocolumn[
\icmltitle{Video Self-Distillation for Single-Image Encoders: A Step Toward Physically Plausible Perception }




\begin{icmlauthorlist}
\icmlauthor{Marcel Simon}{comp}
\icmlauthor{Tae-Ho Kim}{comp}
\icmlauthor{Seul-Ki Yeom}{comp}
\end{icmlauthorlist}

\icmlcorrespondingauthor{Seul-Ki Yeom}{skyeom@nota.ai}

\icmlaffiliation{comp}{Nota AI GmbH, Friedrichstrasse 200, 10117 Berlin, Germany}

\icmlkeywords{self-supervised image-pretraining}

\vskip 0.3in
]



\printAffiliationsAndNotice{}  

\begin{abstract}
Self-supervised image encoders such as DINO \cite{caron2021dino} have recently gained significant interest for learning robust visual features without labels. However, most SSL methods train on static images and miss the temporal cues inherent in videos. We introduce a video-distilled \emph{single-image} encoder trained to predict the next-frame representation from the current frame. This simple objective injects 3D spatial and temporal priors without optical flow or tracking. When pre-training on a single 2-hour video, our approach raises the mean Intersection‑over‑Union (mIoU) on ADE20K from 35.0 (DoRA \cite{venkataramanan2024dora}) to 36.4 while remaining a drop-in replacement for image-only pipelines. Our results highlight video self-distillation as a lightweight route to geometry-aware perception—an essential ingredient for physically plausible world models and \textit{Physical AI}.
\end{abstract}

\section{Introduction and Related Work}\label{sec:intro}

Self-supervised learning (SSL) has reached parity with full supervision for image encoders, powered by self-distillation~\cite{grill2020byol,caron2021dino,oquab2023dinov2}, contrastive objectives~\cite{chen2020simclr,he2020moco}, and masked reconstruction~\cite{he2022mae}.  
Yet these \emph{static-image} paradigms overlook the temporal coherence and multi-view geometry that videos offer for free. Consecutive frames provide virtually unlimited dense supervision and better match downstream settings such as robotic control in \textit{Physical AI}.

Early video-SSL work produced purely video backbones—e.g.\ TimeSformer~\cite{bertasius2021timesformer} and MaskFeat~\cite{wei2022maskfeat}.  More recently, distilling \emph{single-image} encoders from videos has drawn attention.  DoRA~\cite{venkataramanan2024dora} tracks objects to create masked images used as additional data for self-distillation, but still optimizes each frame independently, limiting temporal reasoning.  PooDLe~\cite{wang2024poodle} enforces equivariance under optical‑flow warps, but flow estimation is slow, brittle, and fails under occlusion.

\textbf{Our approach.} 
We address these issues with a training-time change:  a lightweight predictor head regresses the teacher representation of frame $t{+}\Delta$ from the student encoding of frame $t$.  The head is discarded afterward, so inference remains fast while the static-image backbone gains 3D and temporal priors.  When pre-training on a single 2-hour video, our approach surpasses previous work in semantic segmentation with less training time and a simpler architecture.

3D-aware features are especially valuable for embodied agents.  Current Vision–Language–Action (VLA) systems such as GR00T N1~\cite{brohan2025gr00t} and OpenVLA~\cite{kim2024openvla} rely on encoders trained on static images; our dense, temporally informed representations supply complementary cues that might unlock more reliable physical reasoning.

\paragraph{Contributions.}
\begin{itemize}
  \item \textbf{Next-frame objective:} We introduce a next-frame objective that injects temporal priors into an off-the-shelf static-image encoder%
  \item \textbf{Dense prediction head:} We design a lightweight dense-prediction head only used at training time.%
\end{itemize}

\section{Proposed Method}
\label{sec:method}
\paragraph{Problem setup.}
Given an unlabeled video $V=\{x_{1},\ldots,x_{T}\}$ we form clips
$\mathcal{C}_{t}=\{x_{t+i\Delta}\}_{i=0}^{K-1}$, where $\Delta$ is a stride
hyper-parameter (default $\Delta{=}30$ frames) and $K{=}3$.  
We apply a pre-crop and then mostly follow \cite{caron2021dino} for each frame, but apply the same pre-crop and global crop to all frames to allow dense prediction. 
Local views are obtained for each frame starting with $x_{t+\Delta}$.

\begin{figure}[!t]
    \centering
    \includegraphics[width=1.0\columnwidth]{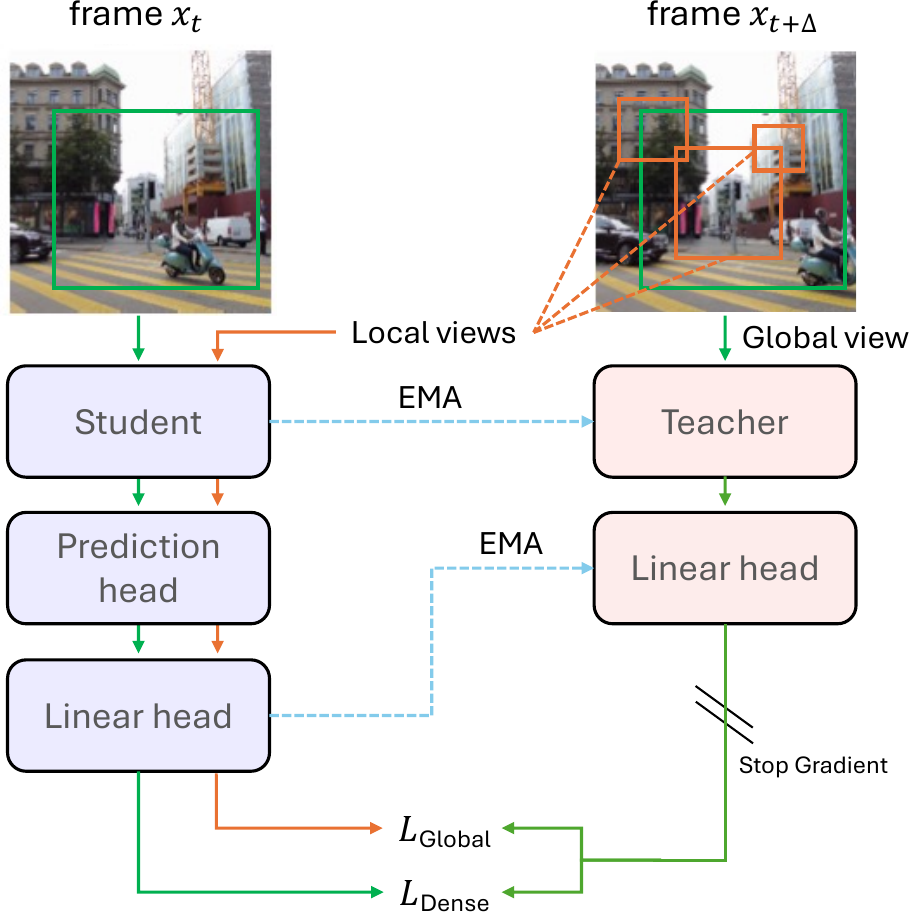}
    \vspace{-1.2em}
    \caption{\textbf{Overview of dense prediction during training.}
    The student encodes a frame and a dense prediction head is used to estimate the features of the teacher for the next frame.
    Exponential moving average (EMA) is used to update the teacher.
    }
    \label{fig:pipeline}
\end{figure}

\paragraph{Architecture.}
Our architecture follows the self-distillation framework presented in \cite{caron2021dino}.
Both teacher and student share a ViT-S backbone as well as the projection head. 
In contrast to DINO, we add an additional 2-layer MLP as well as two attention blocks between the student backbone and projection head.
The attention blocks provide predicting capabilities, while the MLP separates the shared image encoder from these prediction blocks.
This prediction head is only used during training and also exclusive to the student.
Teacher weights are an exponential moving average of the student, identical to DINO, and no gradients flow through the teacher.

\paragraph{Loss functions.}
During training, the student processes all global views except the last one as well as all local views.
The teacher processes only the global views starting with $x_{t+2\Delta}$.
We then optimize two complementary terms. 

\noindent\textbf{(i) Dense next-frame loss.} 

Let $P$ be the number of patch tokens.  
For every consecutive pair $(x_{j},x_{j+\Delta})$ with
$j\in\{t,\ldots,t+(K\!-\!2)\Delta\}$,
the teacher T produces patch tokens $z^{\mathrm{T}}_{j+\Delta}\in\mathbb{R}^{P\times d}$
while the student S predicts
$\hat z^{\mathrm{S}}_{j\!\rightarrow\!j+\Delta}$ from $x_{j}$.
We compute the averaged cross-entropy for each such pair and each patch 
\begin{equation}
\mathcal{L}_{\text{dense}}=
\frac{1}{K-1}\!
\sum_{j=t}^{t+(K-2)\Delta}\;
\operatorname{P-CE}\!\bigl(
      \sigma_{\tau_{\text{T}}}(z^{\mathrm{T}}_{j+\Delta}),
      \sigma_{\tau_{\text{S}}}(\hat z^{\mathrm{S}}_{j\!\rightarrow\!j+\Delta})
\bigr),
\label{eq:dense_new}
\end{equation}  

where $\operatorname{P-CE}$ denotes the averaged per-patch cross-entropy to make notation easier to read.

\noindent\textbf{(ii) Global loss.}  
Each future frame $x_{j}$ with
$j\in\{t+\Delta,\ldots,t+(K-1)\Delta\}$ has $L$ local crops.
Denote its teacher [CLS] token output by $\bar z^{\mathrm{T}}_{j}$ and the $\,\ell$-th
student prediction by $\tilde z^{\mathrm{S}}_{j,\ell}\,$ for $\ell\!=\!1,\ldots,L$.
The loss
\begin{equation}
\mathcal{L}_{\text{global}}=
\frac{1}{(K-1)L}\,
\sum_{j=t+\Delta}^{t+(K-1)\Delta}
\sum_{\ell=1}^{L}
\operatorname{CE}\!\bigl(
      \sigma_{\tau_{\text{T}}}(\bar z^{\mathrm{T}}_{j}),
      \sigma_{\tau_{\text{S}}}(\tilde z^{\mathrm{S}}_{j,\ell})
\bigr).
\label{eq:global_new}
\end{equation}
encourages feature consistency across different augmentations of the same image.

\noindent\textbf{Total objective.}  
The final training loss is the unweighted average
\(
\mathcal{L}=0.5\,\mathcal{L}_{\text{dense}}+0.5\,\mathcal{L}_{\text{global}}.
\)
While the loss computation is similar to DINO, the features used are different. 
The global loss is computed using only the \emph{[CLS] tokens} of views of the same frame.
In contrast, the dense loss is computed using the \emph{patch tokens} of consecutive frames.

\paragraph{Why prediction rather than reconstruction?}
Regressing teacher features teaches the student to preserve only those signals that remain stable over a temporal offset~$\Delta$, implicitly suppressing fleeting noise (e.g., specular flicker) and promoting geometrically consistent embeddings.  
The resulting encoder is a drop-in replacement for single-image pipelines—unlike multi-frame backbones—and is therefore directly usable in VLA-based architectures for robotics and other time-critical settings that demand real-time, physically plausible observations.

\section{Experiments}\label{sec:experiments}

We evaluate the quality of our approach by evaluating the trained static-image encoder on dense prediction tasks.

\subsection{Experimental Setup}

\textbf{Data and sampling.}  
All models are trained from scratch on the publicly available \emph{Walking Tours Venice} 60 fps video (1920\,$\times$\,1080) \cite{venkataramanan2024dora}.  
Clips of $K{=}3$ consecutive frames
$\{x_{t},x_{t+\Delta},x_{t+2\Delta}\}$ are drawn with stride
$\Delta\!=\!30$ unless stated otherwise.

\textbf{Augmentations.}  
Each frame receives a pre-crop with target area $0.05$–$0.20$ to approximate ImageNet statistics, followed by DINO
multi-crop \cite{caron2021dino}.  
The same \emph{global} 224² crop is applied to all three frames, but the color augmentations are independent.
We sample five 96² \emph{local} crops from $x_{t+\Delta}$ and $x_{t+2\Delta}$ each.

\textbf{Training.}  
We train a ViT-S/16 backbone on four NVIDIA RTX 4090 GPUs with an aggregate batch size of 256. 
Running 100 epochs on the \emph{WT Venice} video finishes in roughly one day.  
All frames are pre-extracted as JPEGs to bypass runtime decoding.  
The remaining hyperparameters and schedules match those of \cite{caron2021dino}.

\textbf{Evaluation.}  
Semantic segmentation is measured on ADE20K \cite{ade20k_dataset} with
full UperNet fine-tuning and a
\emph{fast linear} probe (Fast-LP) with frozen backbone, 1000 iterations, and batch size 64 for ablations.
Object detection uses MS COCO-2017 \cite{coco_dataset}.  
The iBOT evaluation protocol and code \cite{zhou2022ibot} is used to provide comparable results to \cite{venkataramanan2024dora}.
Teacher weights are used at test time.
We retrain all baselines ourselves, but use the official DoRA checkpoint for 100 epochs on WT Venice due to excessive training time when using the official codebase.

\subsection{Main Results}

\begin{table}[t]
\caption{Semantic segmentation on ADE20K (mIoU) when pre-trained on WT Venice for 100 epochs. $^\dagger$~result from \cite{wang2024poodle}. DoRA results are computed using the official checkpoint.}
\vspace{0.5em}
\label{tab:ade20k}
\centering
\small
\setlength{\tabcolsep}{5pt}  
\begin{tabular}{lccc}
\toprule
Method & Backbone & UperNet & Fast-LP \\
\midrule
PooDLe & ResNet-50 & 36.6$^\dagger$ & 14.6$^\dagger$ \\
\midrule 
DINO (frames)     &  ViT-S/16      &  31.7 & 12.9 \\
\quad+ pre-crop   &      &  35.1 & 15.8 \\
\quad+ time-aug $\Delta\!=\!5$ &       & 34.9 & 15.9 \\
DoRA   & ViT-S/16  & 35.0 & 17.0 \\
\midrule
\textbf{Ours (dense+global loss)}       & ViT-S/16 & \textbf{36.4} & \textbf{18.3} \\ 
\quad\textit{\,global-loss only}        &          & 34.5 & 15.6 \\
\quad\textit{\,dense-loss only}         &          & 36.2 & 17.4 \\

\bottomrule
\end{tabular}
\end{table}

\begin{table}[t]
\caption{Object detection on MS-COCO. Training uses a $512\times512$ crop for efficiency reasons.}
\vspace{0.5em}
\label{tab:coco}
\centering\small
\begin{tabular}{l c}
\toprule
Method & COCO-2017 mAP \\
\midrule
DINO on frames + pre-crop & 33.3\\
\quad+ time-aug $\Delta{=}5$ & 33.3\\
DoRA & 33.0\\
\midrule
\textbf{Ours} ($\Delta{=}30$) & \textbf{33.5}\\
\bottomrule
\end{tabular}
\end{table}

\paragraph{Semantic segmentation} We present the results for semantic segmentation in Table~\ref{tab:ade20k}.
Our approach improves over DoRA by $+1.4$ mIoU on UperNet and by $+1.1$ mIoU on the
fast linear probe.
We also compare to several DINO baselines.
When using plain DINO and treating the frames of the video as independent images, the accuracy drops by up to $5.2$ reaching only $12.9$\,mIoU on the linear probe.
Pre-cropping proves to be fundamental and improves this baseline by almost $3$ points.
This indicates that the video frames of WT Venice contain too many different objects when using a single global objective.
The last two rows present the mIoU when using only one of the two loss terms and the combined loss achieves the best result.

We also provide a time-augmentation baseline for comparison.
In this case, we use regular DINO but global and local views are split among two consecutive frames instead of using the same frame for all views.
Smaller strides give better results for this baseline, so we use only $\Delta=5$.
As can be seen, the time-based augmentation alone does not give the same benefit as our approach.

Finally, we also include the results from \cite{wang2024poodle}, a recent publication that uses optical flow to compute a dense and global loss across consecutive frames.
Our approach does not require any optical flow and still achieves comparable results.

\paragraph{Object detection}
Table~\ref{tab:coco} shows results for object detection on MS COCO.
Our approach slightly improves the results of DoRA by $0.5$\,mAP while the training time of our approach is significantly faster.

\subsection{Stride Ablation}

\begin{figure}[t]
\centering
\begin{tikzpicture}
\begin{axis}[
    width=\linewidth, height=4.4cm,
    xlabel={Stride $\Delta$ (frames)}, ylabel={Linear mIoU (\%)},
    xmin=-2, xmax=62, xtick={0,3,5,10,30,60},
    ymin=17.5, ymax=18.5, ytick={17.5,18.0,18.5},
    grid=both, grid style={gray!15},
    mark size=2.2pt, legend style={draw=none,font=\footnotesize},
]
\addplot+[thick,mark=o]
coordinates {
    (1, 17.8)
    (3, 17.87)
    (5, 17.99)
    (10, 18.08)
    (30, 18.26)
    (60, 17.72)
};
\end{axis}
\end{tikzpicture}
\vspace{-0.8em}
\caption{Effect of prediction stride $\Delta$ on ADE20K fast-linear accuracy.  
Performance peaks at $\Delta\!\approx\!30$ and plateaus for longer horizons.}
\label{fig:stride}
\end{figure}
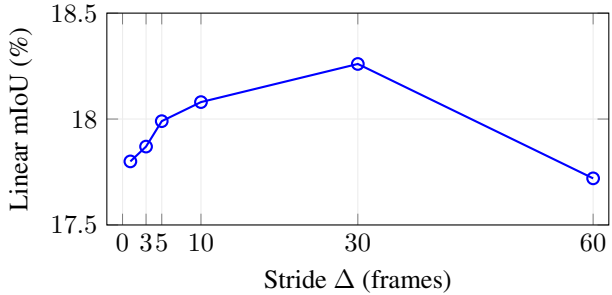

Fig.~\ref{fig:stride} presents results for different values of $\Delta$.
Longer strides modestly improve segmentation up to
$\Delta\!=\!30$, after which gains saturate, confirming that our approach benefits
from extended temporal context, but even works in a frame-by-frame regime.

\section{Conclusion}\label{sec:conclusion}

We propose a lightweight next-frame distillation framework that injects temporal and geometric priors into an off-the-shelf image encoder while preserving single-frame inference speed. Our approach outperforms the tracking-based DoRA on ADE20K and COCO, illustrating that video self-distillation can serve as a pragmatic building block for the \emph{physically plausible world models} sought by Physical AI models. 
Our approach does not require any labels and hence can be directly applied and trained on application specific tasks where sample video data is available. 

\section{Acknowledgments}
This work was supported by the Technology Innovation Program (RS-2024-00468747, Development of AI and Lightweight Technology for Embedding Multisensory Intelligence Modules) funded by the Ministry of Trade Industry \& Energy (MOTIE, Korea).

\bibliography{main}
\bibliographystyle{icml2025}

\end{document}